\documentclass{article}

\PassOptionsToPackage{numbers, compress}{natbib}
\usepackage[preprint]{neurips_2021}

\usepackage[utf8]{inputenc} 
\usepackage[T1]{fontenc}    
\usepackage{hyperref}       
\usepackage{url}            
\usepackage{booktabs}       
\usepackage{amsfonts}       
\usepackage{nicefrac}       
\usepackage{microtype}      
\usepackage{xcolor}         

\usepackage{times}
\usepackage{latexsym}

\usepackage{xspace}
\usepackage{courier}

\def\ie{\emph{i.e.,~}}
\def\eg{\emph{e.g.,~}}

\newcommand{\appname}{\texttt{CausalNLP}\xspace}
\newcommand{\appurl}{https://github.com/amaiya/causalnlp}
\usepackage{amsmath}
\usepackage{amssymb}
\usepackage{amsfonts}
\usepackage{listings}
\usepackage{color}
\usepackage{textcomp}
\definecolor{listinggray}{gray}{0.9}
\definecolor{lbcolor}{rgb}{0.9,0.9,0.9}
\usepackage{balance}

\newcommand\given[1][]{\:#1\vert\:}

\title{CausalNLP: A Practical Toolkit for \\ Causal Inference with Text }

\author{%
  Arun S. Maiya \\
  Institute for Defense Analyses (IDA)\\
  \texttt{amaiya@ida.org} \\
}

\begin{document}

\maketitle

\begin{abstract}
Causal inference is the process of estimating the effect or impact of a treatment on an outcome with other covariates as potential confounders (and mediators) that may need to be controlled.  The vast majority of existing methods and systems for causal inference assume that all variables under consideration are categorical or numerical (\eg gender, price, enrollment).   In this paper, we present \appname, a toolkit for inferring causality with observational data that includes {\em text} in addition to traditional numerical and categorical variables. \appname employs the use of {\em meta-learners} for treatment effect estimation and supports using raw text and its linguistic properties as a treatment, an outcome, or a ``controlled-for'' variable (\eg confounder).  The library is open-source and available at: \url{\appurl}.
\end{abstract}

\section{Introduction and Motivation}

In this work, we present a simple but effective approach to inferring causality from observational data that includes {\em text}. Our approach blends well-studied methods for causal inference from econometrics and statistics with those from natural language processing (NLP).  Indeed, causal questions in social science can often include a text component, which has historically been underutilized \cite{athey2019machine,roberts2020adjusting}. For example, surveys often include open-ended questions with responses in the form of unstructured text. News stories and social media have associations with (and potential causal impacts on) beliefs and attitudes (\eg \citep{romer2021patterns}).  Text and its linguistic properties (\eg topic, sentiment, emotion, readability, politeness, toxicity) can represent both a treatment variable or covariates for which adjustments are needed for unbiased causal estimates.\footnote{A treatment is an independent variable that potentially {\em causes} some outcome, response, or effect. (We will use the terms outcome, response, and treatment effect interchangeably when referring to the dependent variable of interest in this paper.)  Confounders (and mediators) are non-treatment independent variables (covariates) that can bias causal estimates unless they are controlled.} Here, we describe some examples to better illustrate this.
\\~
\\~
\noindent
{\bf Example 1: Text as Treatment or Outcome.}  How does the readability of an email soliciting donations affect whether or not a donation is made?  Or, what is the impact of politeness in an email on receiving fast responses from customer service \citep{egami2018make}?  Both of these questions involve text as a potential {\bf cause} of some outcome. Such linguistic properties of text can also be considered as an {\bf outcome} in a study (\eg emotion of a response in a debate transcript).
\\~
\\~
\noindent
{\bf Example 2: Controlling for Text.}  What is the impact of including a theorem in a research paper on the paper's acceptance \citep{veitch2020adapting}?  Papers on certain topics may have more (or less) theorems and these topics may also impact a paper's acceptance. Thus, the text is a {\em confounding} variable that must be controlled to estimate the impact of inclusion of theorems \citep{mackinnon2000equivalence}. Text may also act as a {\em mediator} that can be controlled in a study \cite{mackinnon2000equivalence}.  For instance, does being male increase the probability of a post being liked or upvoted?  Gender is simply a categorical variable. However, aside from gender, the text of the post (\eg topic, sentiment, emotion) will obviously also have an impact on whether a post is shared. Suppose women tend to write more positive posts or post about popular topics.   We would like to be able to estimate the {\em direct} impact of gender on whether a post is upvoted independent of the text.\footnote{Conditioning on mediators must be done carefully to avoid inadvertently introducing collider bias \citep{cinelli2021crash}.}   
~\\
~\\
\noindent
{\bf Contributions.} Despite the prevalence of natural language text and its importance and impact on larger systems as illustrated above, there is surprisingly little work (relative to other areas) on how best to incorporate text when performing causal analyses. In fact, it is only recently that this problem was more formalized as a research question (\eg \citep{gultchin2021operationalizing,pryzant2021causal,roberts2020adjusting,veitch2020adapting}).  As a result, there are few practical tools or libraries that can be leveraged in causal impact studies with observational data including a text component. To this end, we present \appname, the first practical toolkit for performing causal inference with text data. Implemented using Python, \appname is open-source, free to use under a permissive Apache license, and available on GitHub at: \url{\appurl}.  Surprisingly, in a study we present in Section \ref{sec:experiments}, we found that a simple approach based on meta-learners in \appname achieved statistically similar performance as deeper architectures (\eg CausalBert \citep{pryzant2021causal,veitch2020adapting}), while simultaneously offering more flexibility and requiring only a tiny fraction of the training time using only a standard laptop CPU.

\section{Causal Inference with Meta-learners}
{\em Meta-learners} are an abundantly flexible class of techniques for causal inference on both experimental and observational data \citep{chen2020causalml,kunzel2019metalearners}.\footnote{Counterfactual outcomes are {\em potential} outcomes that can never be observed. For example, if you take a medication that later makes you feel better, what would have happened if you had {\em not} taken the medication is the counterfactual outcome.}   The basic idea behind meta-learners is to use underlying machine learning models (called {\em base learners}) to predict counterfactual outcome estimates from the covariates (\ie the auxiliary independent variables that are not the treatment). Armed with these predictions, the meta-learner can compute the treatment effect for each individual observation in a  straightforward manner.  Note that the term ``meta-learner'' holds a different meaning in machine learning than in econometrics.  We are using the econometrics meaning of the term, as described next in more detail.

\subsection{Types of Meta-learners}
Several different types of meta-learners have been proposed in the literature \citep{kunzel2019metalearners}. One of the simplest meta-learning methods is the {\em T-Learner} described by Athey and Imbens \citep{athey2015recursive} and Kunzel et al. \cite{kunzel2019metalearners}. A T-Learner uses two base learners.  The first is a {\em control} model trained on only those  observations that did not receive treatment, and the second is a {\em treatment} model trained on only those observations that {\em did} receive the treatment. Thus, the following response functions are estimated from observations:
$$\mu_0(x) = {\mathbb{E}}[Y (0)|X = x]$$
$$\mu_1(x) = {\mathbb{E}}[Y (1)|X = x]$$

where each $Y_i(0)$ represents the outcome when observation $i$ is {\em not} assigned the treatment, $Y_i(1)$ is the outcome when it is, and $X$ represents the covariates.   Neither the treatment model nor the control model use the treatment variable as a predictor in a T-Learner. With these two models, the meta-learner can produce two estimates for every observation: $\hat{\mu}_1(x)$ and $\hat{\mu}_0(x)$.  The difference of these two estimates, $\hat{\tau}(x)$, is the T-Learner's estimate of the treatment effect: $\hat{\tau}(x) = \hat{\mu}_1(x) - \hat{\mu}_0(x).$

Other types of meta-learners include the S-Learner \citep{kunzel2019metalearners}, X-Learner \citep{kunzel2019metalearners}, and R-Learner \citep{nie2020quasioracle}. For instance, in contrast to the T-Learner, the S-Learner (an equivalently simple approach) {\em does} include the treatment as a feature similar to other features in a single model:
 $$\mu(x, w) = {\mathbb{E}}[Y^{obs}|X = x, W=w],$$ where $W$ is the treatment and $Y^{obs}$ are the observed outcomes. In the S-Learner, the treatment effect estimate is simply: $\hat{\tau}(x) = \hat{\mu}(x,1) - \hat{\mu}(x,0)$, where $\hat{\mu}$ is the fitted estimator.

All aforementioned meta-learners are supported in \appname and can handle confounding under the ignorability assumption. For a more detailed explanation of these meta-learners, please see \citep{kunzel2019metalearners}.

\subsection{Why Use Meta-learners for Text Data?}
There are three key advantages to using meta-learners when inferring causality with text data.  The first is unconstrained model choice.  Meta-learners can be used with virtually any underlying machine learning model as a base learner (\eg neural architectures for text classification, gradient boosting machines, linear regression). \appname currently uses the LightGBM implementation of gradient boosted decision trees as the default base learner \citep{ke2017lightgbm}, but this is easily configurable. The second advantage is the fact that meta-learners naturally estimate {\em heterogeneous} treatment effects \cite{kunzel2019metalearners}. That is, meta-learners can estimate how causal impacts vary across observations.  If observations represent social media posts, for example, individual posts with larger or smaller treatment effects can easily be identified.  Lastly, as described in the next section, it is very easy to use text features in combination with traditional numerical and categorical variables in any meta-learning approach to causal inference.

\section{Using Text Data with Meta-learners}

Like most traditional causal inference methods, meta-learners represent each observation as a fixed-length vector representing the categorical or numerical attributes of the observation (\eg age, blood pressure, gender, occupation, price, ethnicity, months as customer).  \appname follows two approaches to leveraging text data with meta-learners: 1) explicit or implicit {\em vectorization} of raw text and 2) {\em autocoding}. We will begin with text vectorization.

\subsection{Using Raw Text Directly as Controls}

We can incorporate text as additional covariates (\eg confounders) in a meta-learner by transforming the raw text into a fixed-length vector representation either explicitly or implicitly (during training).  

{\bf Explicit Representations.} A simple and straightforward approach is to represent text fields as TF-IDF vectors \cite{rajaraman2011mining}, which can be fed to almost any machine learning model (\eg logistic regression, random forests, multilayer perceptrons, support vector machines). Other available encoding schemes include those based on topic modeling \citep{blei2003latent} and transformer models fine-tuned on a natural language inference task \citep{wolf2020huggingfaces}. Such feature vector representations of text can be easily used in conjunction with any base machine learning model and any number of traditional numerical or categorical covariates representing supplemental attributes of observations (\eg age, occupation, publication source, etc.). For explicit text representations, we focus on simple and quick-to-train approaches in this paper (\eg TF-IDF with gradient boosted trees as base learners).

{\bf Implicit Representations.}  One can also {\em implicitly} vectorize observations by employing a base learner trained end-to-end that transforms raw text and supplemental covariates into outcome predictions (\eg wide and deep neural architectures \citep{cheng2016wide}). Such representations are learned automatically during estimation (or training) of the causal model.  \appname includes an implementation of {\em CausalBert} adapted from \citep{pryzant2021causal,veitch2020adapting}, which is essentially an S-Learner that uses a DistilBERT text classification model \citep{wolf2020huggingfaces} as the base learner with extra covariates representing additional confounders. Moreover, similar to a TarNet estimator \citep{shalit2017estimating}, {\em CausalBert} maintains a separate set of parameters (or weights) for each possible value of the treatment. Each sample is used to update the weights associated with the corresponding observed treatment value, which serves to preserve the effect of treatment in high-dimensional settings \citep{shalit2017estimating}. In other words, this is intended to address the possible bias towards zero in S-Learners, as described by Kunzel et al. \citep{kunzel2019metalearners}.  See Appendix \ref{sec:appendixb} for details.

\subsection{Autocoding Text for Causal Analyses}

In the previous section, we adjusted for text in a general sense without a specific linguistic property in mind.  That is, we leave it to the base learners to discover which linguistic properties are of relevance to the problem through a sparse (TF-IDF) or dense (embedding) vector representation of the text.  In some cases, analysts may be interested in {\em specific} linguistic properties like sentiment, emotion, or topic. For instance, the sentiment of a document may clearly be a potential confounder in a particular study. In other cases, analysts may need to {\em create} a binary treatment variable from raw text to test a causal hypothesis. To test the causal effect of sentiment on sales, for example, analysts must derive a variable to identify which texts are positive and which are not. Such derived linguistic properties can be supplied directly to any meta-learner as controls, treatments, or outcomes.

\begin{table*}
{\small
\centering
\begin{tabular}{lll}
\hline
\textbf{Analyzer} & \textbf{Columns Created} & \textbf{Example Usage}\\
\hline
Sentiment & positive, negative & \texttt{df = ac.code\_sentiment(texts, df)}\\
Emotion & joy, anger, fear, sadness & \texttt{df = ac.code\_emotion(texts, df)} \\
Topic  & (user-specified topics) & \texttt{df = ac.code\_custom\_topics(texts, df, labels)}\\
User-Defined  & (any linguistic property ) &  \texttt{df = ac.code\_callable(texts, df, fn)}\\
\hline
\end{tabular}
\caption{\label{table:autocoder}
{\bf Autocoder.} Examples of built-in text analyzers available in  Autocoder. In the {\bf Example Usage} column, \texttt{ac} is an Autocoder instance, \texttt{texts} is a list of raw texts as Unicode strings, and \texttt{df} is a {\em pandas} DataFrame containing the dataset under analysis.  For the \texttt{code\_callable} method, \texttt{fn} is any Python callable that takes text as input and returns a dictionary with desired column names as keys and numerical values (\eg probabilities,  scores) as values. After invocation, \texttt{df} will contain newly created columns containing the variables  of interest.
}
}
\end{table*}

The term {\em coding} in survey analysis refers to the process of thematically grouping similar responses together, so that the text responses can be analyzed as a conventional categorical variable.  The {\em Autocoder} in \appname is a built-in suite of state-of-the-art text analyzers that transform raw text into either a binary treatment variable or additional numerical or categorical covariates for any causal inference study involving text. Table \ref{table:autocoder} shows examples of out-of-the-box text analyzers currently included in \appname. Each are implemented using state-of-the-art transformer models \citep{wolf2020huggingfaces}.  For instance, the \texttt{code\_custom\_topics} (and \texttt{code\_sentiment}) methods are implemented using {\bf zero-shot text classification} where a transformer model fine-tuned on a natural language inference task is used to detect user-specified topics (or sentiment) {\em without} training examples \citep{maiya2020ktrain,yin2019benchmarking}. Here, we show a self-contained example:
\begin{lstlisting}[language=Python]
# Zero-Shot Topic Classification with Autocoder

# sample data
comments = ["What is your favorite sitcom?", 
            "Can't wait to vote!"]
import pandas as df
df = pd.DataFrame({
    'OVER_18': ['yes', 'no'],
     'COMMENT' : comments,
      })
      
# autocode text in DataFrame
from causalnlp import Autocoder
ac = Autocoder()
labels=['TV', 'POLITICS']
df = ac.code_custom_topics(df['COMMENT'].values, 
                           df, labels)
\end{lstlisting}

After executing the Python code above, the {\em pandas} DataFrame, \texttt{df}, will contain two new columns (\texttt{television} and \texttt{politics}) containing predicted probabilities representing the degree to which the text pertains to each of these topics:

{\small
 \begin{verbatim}
OVER_18  TV     POLITICS  COMMENT
yes	     0.98 	 0.00      Favorite sitcom?       
no       0.00 	 0.95      Can't wait to vote! 
 \end{verbatim}
 }

These new columns can either be used as controls in a causal analysis or used as a treatment (or outcome) with possible binarization.

\section{Experiments}\label{sec:experiments}

In this section, we evaluate the ability of \appname to recover causal effects from data that includes text. Evaluating such causal effect estimates is challenging because ground-truth treatment effects are typically unavailable. Existing works in this area address this problem by generating semi-synthetic datasets where the text is real and the outcomes are simulated \citep{pryzant2021causal,veitch2020adapting}.  
~\\
~\\
\noindent
{\bf Datasets.}  We use semi-simulated datasets of Amazon reviews generated from the same simulator used in \citep{pryzant2021causal}. The task at  hand involves estimating the causal effect of a positive review on whether or not a product is clicked.  The true sentiment of the review is the treatment, as determined by the rating. Confounders include a categorical variable indicating the product type. We assume the rating is hidden and sentiment is inferred from text.  Thus, text plays a role as both a treatment and a ``controlled-for'' variable.  The simulation script is available here.\footnote{\url{https://github.com/rpryzant/causal-text/blob/main/src/simulation.py}} Aside from varying the random seeds for each trial, default settings were used when simulating outcomes. Average Treatment Effect (ATE) is considered.
~\\
~\\ 
\noindent
{\bf Methods.} We evaluate S-Learners, T-Learners, X-Learners, and R-Learners in addition to {\em CausalBert} \citep{veitch2020adapting} (\ie a kind of S-Learner that uses DistilBERT as base learner) and {\sc TextCause}\footnote{\url{https://github.com/rpryzant/causal-text}} \citep{pryzant2021causal}.  All methods except {\sc TextCause} are currently implemented in \appname.\footnote{Source code for \appname is here: \url{\appurl}.}  To create the binary treatment from the raw text (\ie positive sentiment of review), we use our aforementioned \appname~{\em Autocoder} for all methods except {\sc TextCause}, which uses its own algorithm.  To adjust for the raw text for the fast-training meta-learners, we use TF-IDF vectorization. 
~\\
~\\
\noindent
{\bf Results.} As shown in Table \ref{table:results}, TF-IDF-based meta-learners in \appname performed statistically similar to {\em CausalBert} while requiring only a tiny fraction of the training time on a 2.6 GHz i7 CPU and offering more versatility.  This is despite giving {\em CausalBert} and {\sc TextCause} the added advantage of a Titan V GPU. Differences among top scores in bold under the {\bf $\Delta$ from Oracle} column were not statistically significant at $p=0.05$, as determined by a one-way ANOVA and Post Hoc Tukey HSD.  The simple TF-IDF-based meta-learners in \appname also offer more versatility and flexibility in practice by allowing any number of additional categorical or numerical variables as covariates.  This better helps analysts to reduce sources of biased estimates by controlling for them.
~\\
~\\
\noindent
{\bf Reproducibility.} \appname is a {\em low-code} library that enables causal analyses with minimal effort. As such, we include the full code to reproduce top-performing methods (\ie those listed in bold in Table \ref{table:results}) in Appendix \ref{sec:appendixa}.

\section{Heterogeneous Treatment Effects}
As mentioned, one of the key advantages of meta-learners (\eg a simple TF-IDF-based T-Learner using gradient boosted trees as base learners) is their natural ability to estimate how causal impacts vary across observations, which are referred to as heterogeneous treatment effects. Both conditional and individualized treatment effects (ITE) can be easily calculated.  For instance, in the Amazon review dataset considered in our experiments, we can calculate the treatment effect for only those reviews that contain the word ``toddler'' (\ie the conditional average treatment effect or CATE) in \appname as follows (see Appendix \ref{sec:appendixa} for the complete code example):

\begin{lstlisting}[language=Python]
#  Example: CATE for reviews containing the word "toddler"
series = df['text']
cate = cm.estimate_ate(series.str.contains('toddler'))
\end{lstlisting}

\begin{table*}
{\small
\centering
\begin{tabular}{llll}
\hline
\textbf{Method} & \textbf{ATE Estimate} & \textbf{$\Delta$ from Oracle} & \textbf{Training Time}\\
\hline
Oracle (ground truth) & 15.88 & 0.0 & -\\
Naive (no confounding adjustments) & 7.72 & 8.16 & -\\
S-Learner w/ LogisticRegression & 14.26 & {\bf 1.62} & <1 sec. (CPU) \\
T-Learner w/ LGBM, {\scriptsize [num\_leaves=100}]  & 12.90 & {\bf 2.98} &  2 sec. (CPU)\\
X-Learner w/ LGBM, {\scriptsize [num\_leaves=500}]  & 12.58 & 3.30 &  13 sec. (CPU)\\
R-Learner w/ LGBM, {\scriptsize [num\_leaves=500}] & 12.01 & 3.87 & 15 sec. (CPU) \\
CausalBert \citep{veitch2020adapting} (\ie S-Learner w/ DistilBERT)    & 18.51 & {\bf 2.63} & 16 min. (GPU)\\
{\sc TextCause} \citep{pryzant2021causal}    & 10.03 & 5.85 & 24 min. (GPU)\\
\hline
\end{tabular}
\caption{\label{table:results}
{\bf Results.} Simpler meta-learners in \appname achieved statistically similar performance to {\em CausalBert}, as measured by {\bf $\Delta$ from Oracle} (lower is better), while training faster on only a CPU. All values are means across random trials. Differences among top values in bold were not statistically significant.
}
}
\end{table*}

\section{Limitations}
Here, we briefly outline some limitations of this study.  First, due to the scarcity of datasets for causal inference from text, we evaluated methods using random trials from a single semi-simulated dataset. The extent to which results hold for other datasets and settings is presently unclear.  Second, we observed that {\em CausalBert} estimates are highly sensitive to model-specific hyperparameters such as $g\_weight$, $Q\_weight$, and $mlm\_weight$. These hyperparameters were simply set based on results from \citep{pryzant2021causal}, but a better understanding of how to set these hyperparameters for accurate causal estimates is needed. For instance, although the settings employed yielded estimates closer to the ground truth in the experiments, they also consistently yielded overestimates. Finally, {\sc TextCause} performed noticeably worse than expected despite using the well-performing {\em CausalBert} internally \citep{pryzant2021causal}. We suspect that this may be due to either the treatment representation or default hyperparameter settings, but this requires further investigation.

\section{Related Work}
Causal inference with text is a relatively nascent research area, but there has been a recent surge of work in the field. A number of works focus on deriving or discovering {\bf treatments} from text  \citep{fong2016discovery,algaba2020econometrics,pryzant2021causal,wooddoughty2018challenges}. For instance, \citep{wooddoughty2018challenges} examined different errors associated with predicting treatment labels with classifiers.  \citep{fong2016discovery} proposed ways to discover and derive treatments from text corpora.   \citep{pryzant2021causal} proved bounds on biases arising from estimated treatments and proposed a method to handle text-based treatments. There are also a few works that examine text as an {\bf outcome} (\eg \citep{sridhar2019estimating}).  Finally, other works focus more on adjusting for text as {\bf confounders} or {\bf mediators} \citep{mozer2018matching,roberts2020adjusting,veitch2020adapting}. For instance, \citep{mozer2018matching} and \citep{roberts2020adjusting} explored the use of text matching to control for text in a causal analysis. More recently, \citep{veitch2020adapting} proposed {\em CausalBert} to adjust for text in causal inference.  For a comprehensive survey of approaches to control for text, the reader may refer to \citep{keith2020text}.   \appname is the first library to unify these tasks into a practical and versatile toolkit.
\section{Conclusion}

In this paper, we presented \appname, a Python library for causal inference from observational data with text. Based on meta-learners, \appname is a versatile toolkit supporting inclusion of text and its linguistic properties as treatments, outcomes, confounders, or mediators in a causal study.  Future work might include better explainability and characterization of meta-learner performance with respect to text. For example, under what circumstances are sophisticated, deep neural models required? Finally, we intend to add support for additional causal inference methods adapted for text, as this emerging area progresses.


{\footnotesize
\bibliography{causalnlp}
\bibliographystyle{plain}
}
\appendix
\begin{center}
{\Large\textbf{Appendix}}
\end{center}

\section{Source Code}\label{sec:appendixa}
\appname can be installed using the pip command in Python: \texttt{pip install causalnlp}
~\\
~\\
{\bf T-Learner and S-Learner: }
~\\
The code to train the T-Learner (and S-Learner\footnote{The S-Learner used the following as base learner:  {\texttt{LogisticRegression(solver='liblinear', penalty='l1', fit\_intercept=False)}}, which uses the scikit-learn implementation. Note that S-Learners with linear base learners simply reduce to multiple regression, so heterogeneous treatment effects are not recovered.}) from Table \ref{table:results} is shown below.
\begin{lstlisting}[language=Python]
# Low-Code Causal Inference with CausalNLP

# load semi-simulated dataset
import pandas as pd
df = pd.read_csv('data.tsv', sep='\t', 
                 error_bad_lines=False)
             
# use Autocoder to create treatment from text
from causalnlp import Autocoder
ac = Autocoder()
df = ac.code_sentiment(df['text'].values, 
                       df, 
                       batch_size=16)
df['T_ac'] = (df['positive'] > 0.5).astype('int')

# fit T-Learner for causal-inference using a tuned base learner
from lightgbm import LGBMClassifier
base_learner = LGBMClassifier(num_leaves= 100, 
                              min_child_weight= 100.0,                      
                              colsample_bytree = 0.59, 
                              min_child_samples = 59, reg_alpha= 10, 
                              reg_lambda= 100, sub_sample=0.64)                       
from causalnlp import CausalInferenceModel
cm = CausalInferenceModel(df, 
                          method='t-learner', # or use s-learner, etc.
                          treatment_col='T_ac', 
                          outcome_col='Y_sim', 
                          text_col='text',
                          include_cols=['C_true'],
                          learner=base_learner)
cm.fit()
ate = cm.estimate_ate() # obtain ATE
\end{lstlisting}

In the code above, \texttt{data.tsv} is a semi-simulated dataset created using this script.\footnote{\url{https://github.com/rpryzant/causal-text/blob/main/src/simulation.py}}  When simulating outcomes, default settings of the simulation script were used (\eg higher levels of confounding). Hyperparameter settings  shown for these best-performing models were found automatically through a randomized search optimized to best predict the observed outcome, \texttt{Y\_sim}.
~\\
~\\
{\bf CausalBert:}
~\\
Using a DataFrame similar to above as input, {\em CausalBert} can be trained as follows:
\begin{lstlisting}[language=Python]
from causalnlp.core.causalbert import CausalBertModel
cb = CausalBertModel(batch_size=8, max_length=128)
cb.train(df['text'], df['C_true'], df['T_ac'], df['Y_sim'])  
ate = cb.estimate_ate(df['C_true'], df['text']) # obtain ATE
\end{lstlisting}

Our implementation is based on the one used by Pryzant et al. \citep{pryzant2021causal} and uses the same hyperparameter settings as defaults (\eg $g\_weight=0.0$, $Q\_weight=0.1$, $mlm\_weight=1.0$).

\section{CausalBert Details}\label{sec:appendixb}
{\em CausalBert}, adapted from \citep{pryzant2021causal} and originally proposed in \citep{veitch2020adapting}, estimates the causal effect, $\tau(W,C)$, using the text $W$, additional covariates $C$, treatment $T$, and outcomes $Y$. It can be viewed as an S-Learner using a DistilBERT text classification (or regression) model \citep{wolf2020huggingfaces} as the base learner. During estimation, the DistilBERT base learner transforms the raw text into a learned vector representation, $\mathbf{b}(W)$, of the confounding information in the texts \citep{pryzant2021causal}.  More specifically, the estimator is trained for the expected conditional outcome $Q(t, \mathbf{b}(W), C) = \mathbb{E}[Y \given T=t, \mathbf{b}(W), C]$:

$$\hat{Q}(t, \mathbf{b}(W), C) = \sigma (\mathbf{M}^b_t \mathbf{b}(W) + \mathbf{M}^c_t \mathbf{c} + b ),$$ 
where the vector $\mathbf{c}$ is a one-hot encoding of the covariates $C$, the vectors $\mathbf{M}^b_t \in \mathbb{R}^{768}$ and $\mathbf{M}^c_t \in \mathbb{R}^{\vert C \vert}$ are learned, one for each  value $t$ of the treatment, and the scalar $b$ is a bias term \citep{pryzant2021causal}. An improvement would be to learn an embedding of $C$ (for covariates that are categorical), but the current experimental implementation of {\em CausalBert} only uses a single one-hot-encoded vector.  

Similar to a TarNet estimator \citep{shalit2017estimating}, the parameters $\mathbf{M}_t$ are updated on examples where $T=t$.  This prevents the treatment effect (and additional confounding information in $C$) from getting lost in the presence of the higher dimensional learned representation, $\mathbf{b}(W)$.  That is, this is intended to prevent the bias towards zero of conventional S-Learners, as described by Kunzel et al. \citep{kunzel2019metalearners}.

As with other S-Learners, once trained, the estimated causal effect, $\hat{\tau}(W,C)$, is:

\begin{align*}
\hat{\tau}(W,C) =&\frac{1}{n}\sum_i \Big[ \hat{Q}(1, \mathbf{b}(W_i), C_i) \nonumber \\
&\quad \quad \quad \quad- \hat{Q}(0, \mathbf{b}(W_i), C_i)  \Big],
\end{align*}

where the expectation over the texts, $W$, and additional covariates, $C$, is approximated with a sample average \citep{pryzant2021causal}.

\end{document}